\documentclass[letterpaper,10pt,conference]{ieeeconf}

\IEEEoverridecommandlockouts
\usepackage[T1]{fontenc}
\usepackage{amsmath,amssymb,amsfonts,mathtools}
\usepackage{graphicx}
\usepackage{booktabs}
\usepackage{cite}
\usepackage{microtype}
\usepackage{capt-of}
\usepackage{balance}
\usepackage[hyphens]{url}
\newcounter{propositioncounter}
\newcounter{corollarycounter}
\newenvironment{plainproposition}[1]{\refstepcounter{propositioncounter}\par\smallskip\noindent Proposition~\thepropositioncounter. #1.\ }{\par\smallskip}
\newenvironment{plaincorollary}[1]{\refstepcounter{corollarycounter}\par\smallskip\noindent Corollary~\thecorollarycounter. #1.\ }{\par\smallskip}
\usepackage{amsthm}
\usepackage[hidelinks]{hyperref}

\title{Banana Kick: Response-Informed Skill Evolution for Humanoid Soccer}

\author{Hao E. Zhang$^{*,1,2}$, Ruize Geng$^{*,1}$, Raihan Haque$^{3}$, Khalil Zbiss$^{3}$, Guanyang Luo$^{3}$, Hui-ping Wang$^{3}$,\\ H. Eric Tseng$^{\dagger,2}$ and Ding Zhao$^{\dagger,1}$
\thanks{$^{*}$Equal contribution.}
\thanks{$^{\dagger}$Correspondance to Ding Zhao (dingzhao@cmu.edu) and H. Eric Tseng (hongtei.tseng@uta.edu).}
\thanks{$^{1}$Hao E. Zhang, Ruize Geng and Ding Zhao are with Carnegie Mellon University.
        (email: haoz4@andrew.cmu.edu; rgeng3@jh.edu; dingzhao@cmu.edu)}%
\thanks{$^{2}$Hao E. Zhang and H. Eric Tseng are with the University of Texas at Arlington.
        (email: haoz4@andrew.cmu.edu; hongtei.tseng@uta.edu)}%
\thanks{$^{3}$Raihan Haque, Khalil Zbiss, Guanyang Luo and Hui-ping Wang are with General Motors.
        (email: raihan.haque@gm.com; khalil.zbiss@gm.com; guanyang.luo@gm.com; hui-ping.wang@gm.com)}%
}

\makeatletter
\let\@oldmaketitle\@maketitle
\renewcommand{\@maketitle}{\@oldmaketitle\centering
  \includegraphics[width=0.99\textwidth]{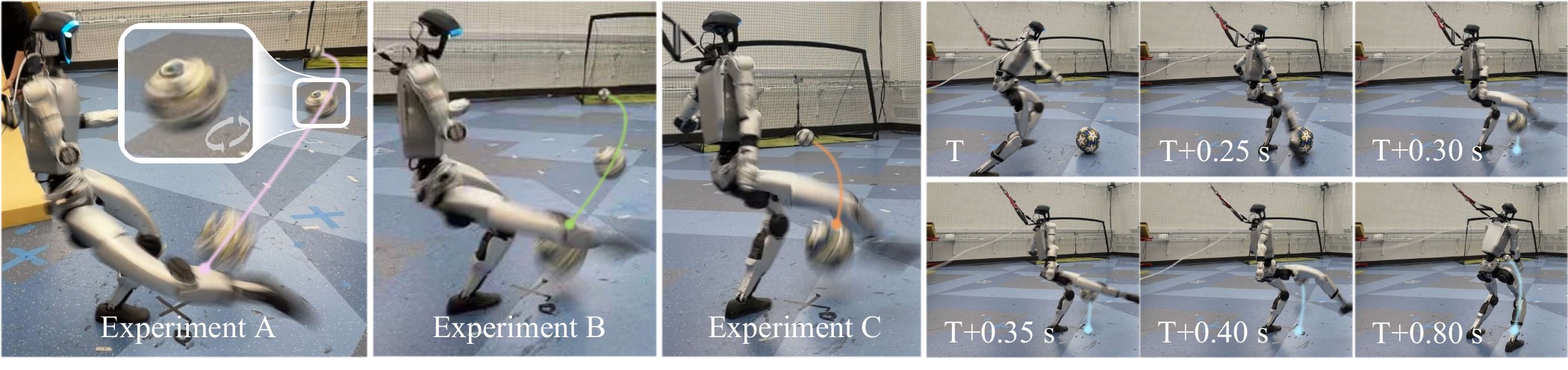}
\captionof{figure}{Representative hardware executions from 30 physical trials show repeatable lateral bending of the RISE-learned banana kick, while the enlarged sequence highlights ball spin underlying the curved flight. Quantitative sim-to-real trajectories are reported in Fig.~\ref{fig:flight}.}
  \label{fig:teaser}
  \bigskip}
\makeatother

\begin{document}

\maketitle

\setcounter{figure}{1} 

\thispagestyle{empty}
\pagestyle{empty}

\begin{abstract}
Humanoid kicking requires coordinated whole-body motion and precise contact, while a banana kick demands contact mechanics that generate ball spin and aerodynamic curvature. Motion imitation provides a reliable ordinary-kick prior, but reinforcement learning may improve shot speed and placement accuracy without changing the underlying kicking technique. Adapting this prior to a qualitatively different contact-rich skill can fail even when the reward is dense and optimization remains stable. The failure occurs when the task objective is locally flat over the current policy's responses. We term this condition first-order learning starvation. To address it, we propose response-informed skill evolution (RISE), a closed-loop objective-continuation method for policy adaptation. RISE ranks bounded objective changes using response sensitivity estimated from cached rollouts and accepts updates only when they produce verified response progress while preserving kicking reliability. Our analysis shows that rescaling a saturated spin reward cannot recover first-order sensitivity at zero spin, whereas adapting coupled contact responses can provide a learnable path to spin generation. We integrate RISE into a humanoid kicking pipeline under calibrated contact and Magnus-force aerodynamics, and sim-to-real transfer. Experiments show that RISE evolves the ordinary kick into a high-spin curved kick with 11.55~rad/s mean ball spin, improves the mean evaluation score by 19.8\% over a learning-progress curriculum, and raises joint target attainment from 15.2\% to 50.9\%. Ablations and response diagnostics support the mechanism, while 30 motion-capture-recorded physical trials demonstrate consistent hardware transfer of the learned curved kick. Project website: \url{https://haozhang-thu.github.io/bananakick/}
\end{abstract}

\section{Introduction}
\label{sec:intro}

Humanoid soccer compresses whole-body coordination and precise contact into a brief, forceful interaction with the ball. A banana kick is harder because the robot must generate substantial spin while preserving balance, ball speed, and placement. Motion imitation provides reliable dynamic priors \cite{peng2018deepmimic}. Physics-aligned sim-to-real training has enabled agile humanoid skills \cite{zhang2026halo}, while reinforcement learning has produced capable robotic soccer behaviors \cite{haarnoja2023soccer}. Yet adapting an ordinary kick into a qualitatively different contact-rich skill is not equivalent to refining speed or accuracy: the policy must leave the prior's familiar contact regime and discover an impact strategy that changes the ball's rotational and translational motion together.

A strong ordinary-kick prior avoids rediscovering balance and swing coordination, but also concentrates exploration around familiar contacts. Spin can require coordinated changes in contact location, tangential foot velocity, and impact timing \cite{bengio2009curriculum}. Even with dense reward and stable optimization, the policy may remain near a reliable straight-kick solution if the objective varies little over its current responses. Accurate simulation is critical here: impact mechanics set outgoing velocity and spin, while aerodynamic forces turn spin into flight curvature \cite{eysenbach2019diayn}.

Prior methods address parts of this problem. Residual reinforcement learning adds learned corrections to nominal controllers \cite{zhang2025bi}; residual skill policies extend this idea to adaptable skill spaces \cite{rana2023residual}. Reward shaping changes the objective \cite{ng1999policy}. Curriculum learning introduces intermediate tasks, while automatic teachers select tasks by competence or learning progress \cite{zhang2026cognition}. Skill-discovery methods broaden behavior through diversity or dynamics-aware objectives \cite{sharma2020dads}, while disagreement-based exploration seeks novel behaviors \cite{portelas2020teacher}. These methods do not directly ask whether the current objective supplies an informative first-order direction over the physical responses induced by a strong motion prior \cite{zhang2026interaction}.

We refer to this mismatch as first-order learning starvation. For an ordinary kick with zero spin, rescaling a saturated spin reward cannot create first-order sensitivity; likewise, a narrow contact-target reward can be nearly flat over the contacts produced by the prior. A sensitive reward on a coupled contact variable can instead move the policy beyond this familiar regime, where spin becomes learnable. This observation motivates response-informed skill evolution (RISE), a closed-loop objective-continuation framework for policy adaptation. Here, skill evolution denotes moving beyond performance refinement by leaving the contact regime encoded by a competent prior and acquiring qualitatively different physical responses. RISE uses measured contact and ball-motion responses to evaluate bounded objective changes and warm-starts policy optimization from the last accepted checkpoint. Because reward sensitivity alone does not guarantee useful skill progress, a candidate is retained only when the resulting policy advances the desired physical responses while preserving correct-foot reliability. Repeating this process provides a continuation path from the ordinary-kick prior toward a new contact regime.

The main contributions are summarized as follows:
1) we demonstrate humanoid skill evolution beyond performance refinement, evolving a reliable ordinary-kick prior into a spin-generating curved-kick skill through a qualitatively different contact regime, with consistent reproduction on real hardware. To the best of our knowledge, this work provides a pioneering physical humanoid demonstration of curved soccer kicking, or banana kick;
2) RISE provides a closed-loop objective-continuation framework that adapts the task objective to the policy's evolving reachable responses and verifies candidate updates through physical progress and kicking reliability;
3) a response-space characterization of first-order learning starvation explains why locally flat objectives can stall prior-based adaptation and how coupled contact responses can restore a viable learning direction. Representative hardware executions are shown in Fig.~\ref{fig:teaser}.

\section{Related work}
\label{sec:related}

\subsection{Curriculum learning and reward design}

Curriculum learning (CL) organizes training through intermediate problems \cite{bengio2009curriculum}, while reverse curricula expand the initial-state distribution from states near the goal \cite{ng1999policy}. Automatic teachers choose tasks from competence or learning progress \cite{portelas2020teacher}; TeachMyAgent benchmarks such strategies across environments \cite{romac2021teachmyagent}. Reward shaping changes the learning signal, with potential-based shaping preserving optimal policies under its assumptions \cite{zhang2025multi}. RISE instead adapts reward parameters from local response sensitivity on the current rollout distribution.

\subsection{Policy adaptation and physical skill discovery}

Residual reinforcement learning (RL) adds corrections to nominal controllers \cite{johannink2019residual}, and residual skill policies learn adaptable skill spaces for downstream control \cite{rana2023residual}. Motion priors provide coordinated behavior for later adaptation \cite{peng2018deepmimic}, while physics alignment supports transfer of agile humanoid skills \cite{he2025asap}. Unsupervised skill discovery promotes diversity \cite{eysenbach2019diayn} or dynamics-aware behaviors \cite{sharma2020dads}; disagreement-based exploration seeks novel experience \cite{pathak2019disagreement}, and quality-diversity methods explicitly maintain behavioral repertoires \cite{pugh2016quality}. Humanoid soccer learning has demonstrated agile kicking \cite{haarnoja2023soccer}; our focus is instead the transition from a competent kick to a distinct spin-generating contact regime when the initial objective gives little local guidance.

\subsection{Continuation and sensitivity-guided search}

Numerical continuation follows parameterized solutions through local changes and warm starts \cite{allgower1990}. Classical experimental design selects informative evaluations \cite{kiefer1959optimal}; fractional designs reduce evaluation cost in structured settings \cite{box1961fractional}. Bayesian optimization uses surrogate models to search large spaces, including random embeddings \cite{wang2016rembo}, additive structure \cite{kandasamy2015additive}, sparse subspaces \cite{eriksson2021saasbo}, and adaptive nested subspaces \cite{papenmeier2022baxus}. RISE instead scores candidate rewards directly on cached responses and verifies the selected change through policy training and measured response progress.

\begin{figure*}[!t]
\centering
\includegraphics[width=\textwidth]{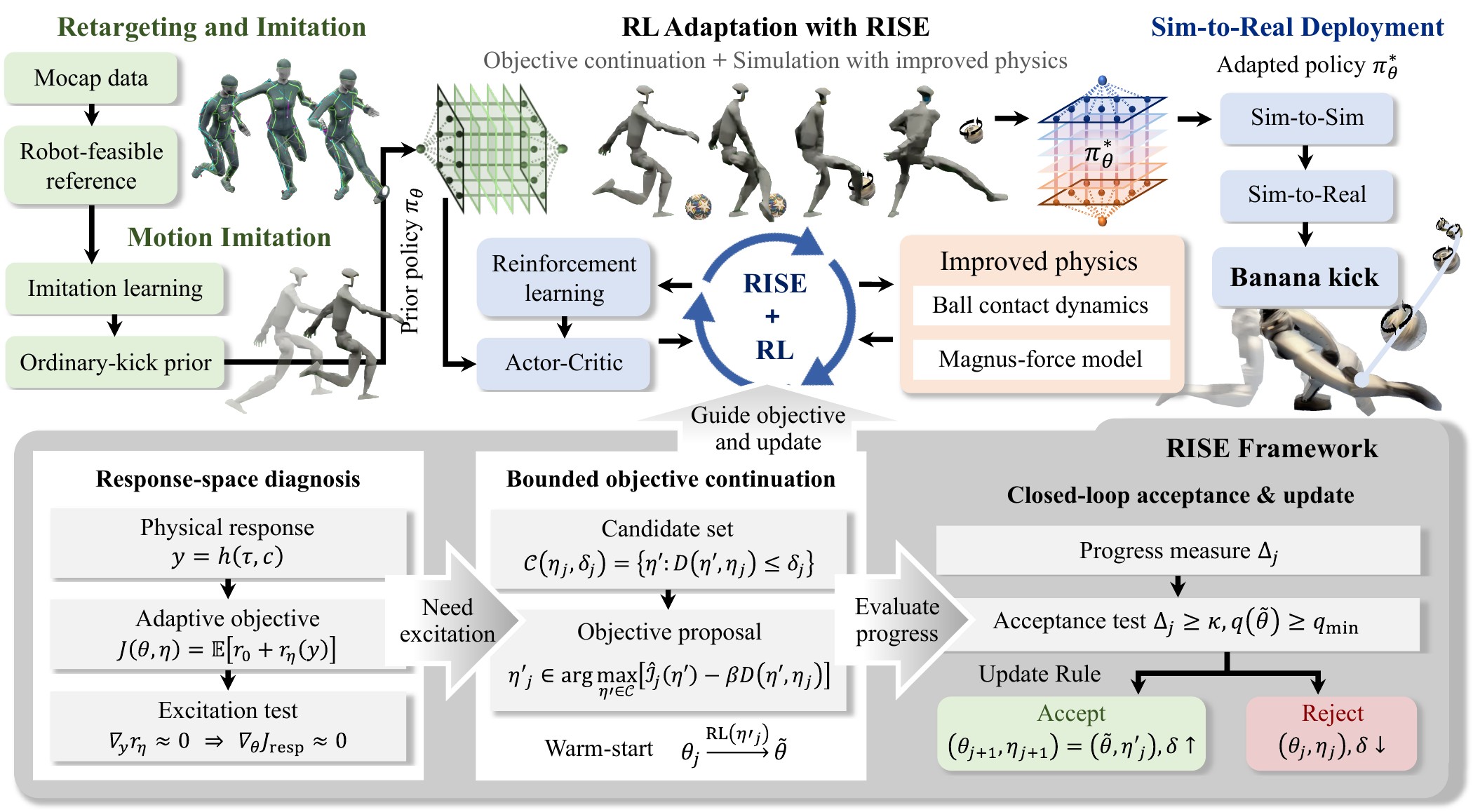}
\caption{From motion prior to new physical skill: imitation provides a reliable ordinary kick while RISE evolves its contact regime under calibrated contact dynamics and Magnus-force aerodynamics. Cached response sensitivities guide bounded objective proposals, and response progress with correct-foot reliability determines whether each policy update is retained before direct hardware transfer.}
\label{fig:method-overview}
\end{figure*}

\section{Methodology}
\label{sec:method}

The architecture of this work, see Fig.~\ref{fig:method-overview}, separates skill initialization from skill evolution. Captured human motion is retargeted to a robot-feasible reference, and imitation learning (IL) provides a coordinated ordinary-kick prior. RISE combined with RL then changes the foot--ball contact regime under calibrated contact and Magnus-force aerodynamics before hardware transfer.

\subsection{Motion capture, retargeting, and imitation}
\label{sec:pipeline}

Motion capture provides a sequence of human kicking poses $\mathcal M=\{z_t^{\mathrm{human}}\}_{t=0}^{T}$. Retargeting accounts for the different human and robot kinematics and maps $\mathcal M$ to a robot-feasible reference trajectory $\bar x_{0:T}=\mathcal T(\mathcal M)$. The reference describes the motion to be tracked, while imitation learns a feedback policy that executes it under simulated dynamics.

The ordinary-kick prior $\pi_{\theta_0}$ is trained with the motion-imitation reward $r_{\mathrm{imit}}(x_t,\bar x_t)$ over the retargeted trajectory. This stage supplies balance and swing coordination for contact adaptation. Let $x_t$ denote the simulator state, $o_t$ the deployable actor observation, and $\pi_\theta(a_t\mid o_t,c)$ the kicking policy for task context $c\sim\mathcal D$. Motion imitation sets $\theta_0$; adaptation updates $\theta$ and the response-reward parameters $\eta$.

\subsection{Reinforcement learning with calibrated physics}
\label{sec:physics-rl}
\label{sec:problem}

The imitation policy initializes curved-kick reinforcement learning. Adaptation changes the foot--ball interaction to generate spin while retaining the learned whole-body coordination. The actor uses deployable proprioception, motion-reference cues, ball-relative state, and the previous action,
\begin{equation}
\begin{aligned}
o_t={}&[\mathbf q_t,\dot{\mathbf q}_t,\boldsymbol\omega_t^B,\mathbf g_t^B,
o_t^{\mathrm{ref}},\\
&\mathbf p_{b,t}^B,\mathbf v_{b,t}^B,a_{t-1}],\\
a_t&\sim\pi_\theta(\cdot\mid o_t,c),\qquad a_t\in\mathbb R^{23}.
\end{aligned}
\label{eq:obs-action}
\end{equation}
Here $\mathbf q_t$ and $\dot{\mathbf q}_t$ are the 23 actuated-joint states, $\boldsymbol\omega_t^B$ is base angular velocity, $\mathbf g_t^B$ is projected gravity, $o_t^{\mathrm{ref}}$ contains the reference-motion tracking cues, and $(\mathbf p_{b,t}^B,\mathbf v_{b,t}^B)$ are ball position and velocity in the robot frame. The policy runs at 50~Hz while the physics and PD control run at 200~Hz. The environment includes calibrated elastic ball contact and an aerodynamic model with the Magnus force:
\begin{align}
m_b\dot{\mathbf v}_b={}&m_b\mathbf g+
\mathbf F_{\mathrm{c}}(x,a;\widehat\psi_{\mathrm c})+\mathbf F_{\mathrm{a}}(\mathbf v_b,\boldsymbol\omega_b;\psi_{\mathrm a}),
\label{eq:ball-dynamics}
\end{align}
where $m_b$, $\mathbf v_b$, and $\boldsymbol\omega_b$ denote ball mass, velocity, and angular velocity, and $\mathbf g$ is gravitational acceleration. The contact parameters $\widehat\psi_{\mathrm c}$ are calibrated to the ball's elastic response. Contact forces and torques determine the outgoing velocity and spin, and the aerodynamic force $\mathbf F_{\mathrm a}$ includes the spin-induced lateral force. Together, these models define the simulation transition law $P_\psi$, with $\psi=(\widehat\psi_{\mathrm c},\psi_{\mathrm a})$ fixed throughout policy adaptation. PPO updates the policy from rollouts under $P_\psi$, while RISE selects the response-reward parameters. Thus, contact and aerodynamic models determine the physical consequences of a kick, and response sensitivity determines which reward change is proposed. A rollout $\tau\sim p_\theta(\tau\mid c;P_\psi)$ induces the measurable physical response $y=h(\tau,c)\in\mathcal Y$, with policy-induced distribution $p_\theta(y)$. PPO maximizes a discounted return with fixed per-step robot-control terms and the adaptable terminal response objective,
\begin{align}
J(\theta,\eta)
&=\mathbb E\!\left[\sum_{t=0}^{T-1}\gamma^t r_{0,t}
+r_\eta(y)\right],\nonumber\\
r_\eta(y)
&=\sum_{k=1}^{K}w_k\phi_k(y_k;t_k,\sigma_k).
\label{eq:objective}
\end{align}
The fixed reward $r_{0,t}$ contains motion tracking, body-stability, action-rate, joint-velocity, and torque regularization and is unchanged across RISE and all baselines. The response objective contains the contact and ball-motion channels adapted by RISE. Here $w_k>0$ and $\sigma_k>0$ are the weight and scale of channel $k$, and $t_k$ is its target when required by the kernel. The continuation variable $\eta\in\mathcal E\subset\mathbb R_{>0}^{d}$ collects only the adjustable response weights and scales; the actor observation, action interface, fixed reward terms, task targets, and dynamics remain unchanged. For curved-kick adaptation,
\begin{equation}
\begin{aligned}
y={}&(e_{\mathrm{patch}},v_{\mathrm{tan}},\omega,v_{\mathrm{ball}}),\\
\eta={}&(w_{\mathrm{tan}},w_{\mathrm{spin}},w_{\mathrm{speed}},\sigma_{\mathrm{patch}},\sigma_{\mathrm{tan}},\sigma_{\mathrm{spin}}).
\end{aligned}
\label{eq:task-param}
\end{equation}
The response coordinates are contact-patch error, tangential impact speed, ball spin, and outgoing ball speed. Patch error is measured relative to the desired strike location. Patch and tangential-speed rewards use target-centered bell kernels, while spin and ball-speed rewards use one-sided saturating kernels. Let $\mathcal G\subseteq\mathcal Y$ denote the desired response set and $q(\theta)$ the probability of correct-foot contact. Starting from $\theta_0$, adaptation seeks to increase $\Pr_{p_\theta}(y\in\mathcal G)$ while maintaining $q(\theta)\ge q_{\min}$. RISE varies $\eta$ to maintain reward sensitivity on the changing response distribution. 

\subsection{Response sensitivity and local analysis}
\label{sec:learnability}
\label{sec:theory}

We assess an objective by its sensitivity to the responses visited by the current policy. The two reward kernels are
\begin{align}
\phi_{\mathrm{bell}}(y;t,\sigma)&=\exp\left[-(y-t)^{2}/\sigma^{2}\right],
\label{eq:bell}\\
\phi_{\mathrm{sat}}(y;\sigma)&=1-\exp\left[-y_{+}^{2}/\sigma^{2}\right],\qquad y_{+}=\max(0,y).
\label{eq:sat}
\end{align}
Channel $k$ is first-order starved when $\partial r_\eta/\partial y_k=0$ almost surely under $p_\theta$. This condition depends jointly on the reward parameters and the response distribution: a smooth kernel can be sensitive near its target and effectively flat at the responses produced by the prior. Let $g_k(y_k;\eta)=|\partial\phi_k/\partial y_k|$ denote the response sensitivity of channel $k$. The normalized excitation:
\begin{equation}
\mathcal I(\theta,\eta)=
\mathbb E_{y\sim p_\theta}
\!\left[\sum_{k=1}^{K}\frac{w_k}{Z_k(\theta)}g_k(y_k;\eta)\right],
\label{eq:excitation}
\end{equation}
where $Z_k(\theta)>0$ normalizes derivative units across channels. At accepted iterate $j$, a fixed bank of $N$ contexts produces cached responses $\mathcal B_j=\{y_j^{(n)}\}_{n=1}^{N}$. Candidate excitation is estimated without additional policy training by
\begin{equation}
\widehat{\mathcal I}_j(\eta')=
\frac{1}{N}\sum_{n=1}^{N}\sum_{k=1}^{K}
\frac{w'_k}{Z_{j,k}}g_k(y_{j,k}^{(n)};\eta').
\label{eq:empirical-excitation}
\end{equation}
The batch and normalization constants are held fixed while scoring candidates. Eq.~\ref{eq:empirical-excitation} measures local reward contrast; whether that contrast leads to progress is checked by the acceptance criterion. Changing $\sigma$ can bring a bell kernel's sensitive region closer to the current response distribution. In contrast, $\partial\phi_{\mathrm{sat}}(0;\sigma)/\partial y=0$ for every $\sigma>0$. To interpret response sensitivity, consider a locally differentiable reparameterization $y=F(\theta,c,\xi)$, where $\xi$ represents exogenous randomness. Define $J_{\mathrm{resp}}(\theta,\eta)=\mathbb E[r_\eta(F(\theta,c,\xi))]$.

\begin{plainproposition}{Response-flat objectives suppress response-mediated first-order updates}
\label{prop:flat}
If $\nabla_y r_{\eta}(F(\theta,c,\xi))=0$ almost surely and differentiation can be interchanged with expectation, then $\nabla_{\theta}J_{\mathrm{resp}}(\theta,\eta)=0$.
\end{plainproposition} 

The result follows from
\begin{equation}
\nabla_\theta J_{\mathrm{resp}}=
\mathbb E\!\left[\left(\frac{\partial F}{\partial\theta}\right)^{\!\top}
\nabla_y r_\eta\right].
\label{eq:response-gradient}
\end{equation}
Small response derivatives likewise imply small response-mediated gradients when $\partial F/\partial\theta$ is bounded. This is a local response-space interpretation, not a requirement to differentiate through contact dynamics: fixed control rewards remain active, RISE uses sensitivity only to rank proposals, and sampled responses verify each policy update.

\begin{plainproposition}{Sensitivity-matched scale}
\label{prop:sigmastar}
For a fixed $y\neq t$, $|\partial\phi_{\mathrm{bell}}(y;t,\sigma)/\partial y|$ is maximized over $\sigma>0$ at $\sigma=|y-t|$. We therefore use
\begin{equation}
\sigma_k^{\star}=\mathbb{E}_{p_{\theta}}[|y_k-t_k|]
\label{eq:sigmastar}
\end{equation}
as a response-matched population reference, not a closed-form maximizer of expected sensitivity. Candidate selection uses the measured excitation in Eq.~\ref{eq:empirical-excitation}.
\end{plainproposition}

\begin{plaincorollary}{Zero one-sided responses cannot be opened by self-rescaling}
\label{cor:sat}
For Eq.~\ref{eq:sat}, $\partial\phi_{\mathrm{sat}}/\partial y=0$ at $y=0$ for every $\sigma>0$. If a response is identically zero, changing only its positive weight or scale cannot create local response-space excitation. Exploration or a coupled response term may nevertheless move the policy away from zero.
\end{plaincorollary}

Together, Proposition~1, Proposition~2, and Corollary~1 characterize a concrete local failure mode and its remedy: a saturated zero-spin channel cannot recover a first-order signal by self-rescaling, whereas a coupled contact channel can restore informative gradients and initiate spin. Section~\ref{sec:mechanism} examines this predicted contact-mediated sequence. Bounded objective changes and warm starts further admit a local continuation interpretation: for a twice continuously differentiable local loss with a regular stationary point and nonsingular policy Hessian, the implicit function theorem yields a locally smooth stationary branch as the objective parameters vary, motivating the bounded warm-started updates used by RISE.

\subsection{Objective continuation and policy transfer}
\label{sec:continuation}
\label{sec:transfer}

At accepted iterate $j$, let $(\theta_j,\eta_j,\delta_j)$ denote the policy, objective parameters, and objective trust-region radius. Candidate objectives lie in the admissible domain $\mathcal A(\eta_j)$ and satisfy $D(\eta',\eta_j)\le\delta_j$, where
\begin{equation}
D(\eta',\eta)=\frac{1}{d}\sum_{i=1}^{d}
\left(\log\eta'_i-\log\eta_i\right)^2
\label{eq:deformation}
\end{equation}
measures relative deformation across positive objective parameters. The proposal is
\begin{equation}
\begin{aligned}
\eta'_j\in\arg\max_{\eta'\in\mathcal A(\eta_j)}
&\ \widehat{\mathcal I}_j(\eta')-\beta D(\eta',\eta_j)\\
\text{s.t.}\quad &D(\eta',\eta_j)\le\delta_j ,
\end{aligned}
\label{eq:proposal}
\end{equation}
where $\beta\ge0$ penalizes objective change. We enumerate the discrete candidate set used in the experiments.

\begin{table}[t]
\centering
\caption{Experimental settings.}
\label{tab:exp-config}
\setlength{\tabcolsep}{1.0pt}
\begin{tabular*}{\columnwidth}{@{\extracolsep{\fill}}lclc@{}}
\toprule
Item & Value & Item & Value \\
\midrule
Robot & G1 & Seeds & $5$ \\
Prior & Ordinary & Bank & $4096$ \\
Offset & $\pm5$ cm & Bootstrap & $10$k \\
Actor & Mean & Reliability & Correct-foot \\
\midrule
$\delta_0$ & $.35$ & $\delta$ bounds & $[.02,.60]$ \\
$\gamma_{\uparrow}/\gamma_{\downarrow}$ & $1.5/.5$ &
$\beta/\kappa/q_{\min}$ & $1/.15/.85$ \\
$S_{\rm patch}$ & $\{.02,.06,.10\}$ &
$S_{\rm tan}$ & $\{3,5,6,8\}$ \\
$S_{\rm spin}$ & $\{15,18,20,25,30\}$ &
$W$ & $\{2/3,1,3/2,2\}$ \\
\midrule
$\lambda_p/\lambda_t$ & $1000/300$ &
$\lambda_\omega/\lambda_v$ & $125/100$ \\
$s_p$ & $.06$ & $t_t/s_t$ & $12/5$ \\
$s_\omega/s_v$ & $18/15$ &
$\bar e/\bar v$ & $6{\rm\,cm}/7{\rm\,m\,s^{-1}}$ \\
$\bar\omega/\bar\alpha$ & $10{\rm\,rad\,s^{-1}}/.90$ &
Physical trials & $30$ \\
\bottomrule
\end{tabular*}
\end{table}

\begin{figure}[t]
\centering
\includegraphics[width=\columnwidth]{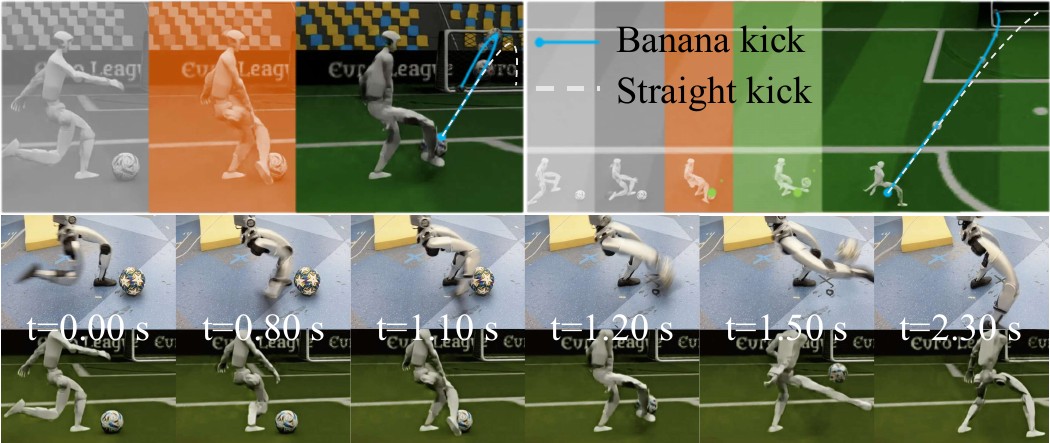}
\caption{Simulation and sim-to-real comparison: (a,b) show the simulated kick and straight-versus-banana trajectories, while (c,d) compare physical and simulated contact-side sweep and follow-through; quantitative flight trajectories are reported in Fig.~\ref{fig:flight}.}
\label{fig:sim-to-real-visual}
\end{figure}

\begin{figure}[t]
\centering
\includegraphics[width=\columnwidth]{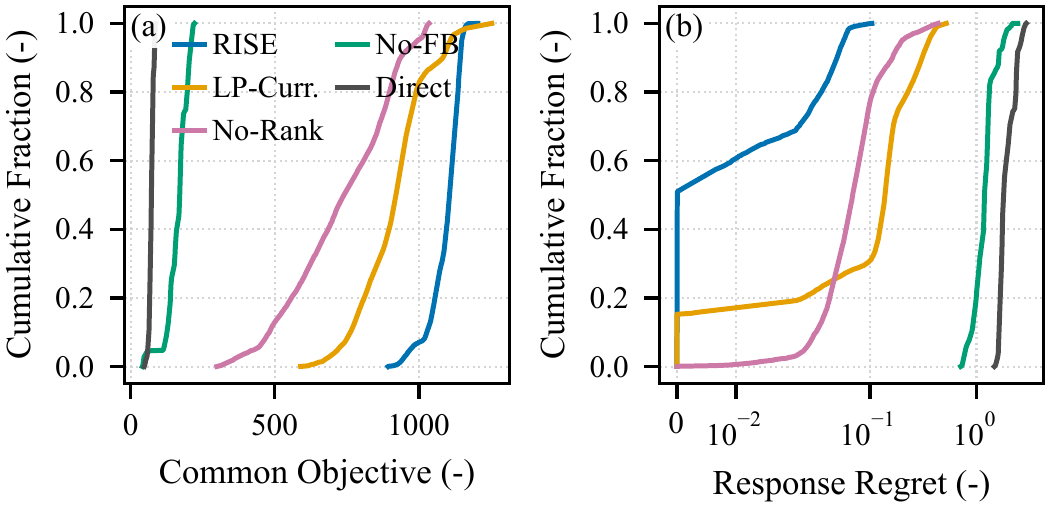}
\caption{Cross-method evaluation over five seeds and 4,096 matched contexts per seed: (a) fixed-objective score and (b) normalized response regret. LP-Curr. denotes the learning-progress curriculum; No-Rank and No-FB remove sensitivity ranking and closed-loop feedback, and Direct denotes direct-final PPO.}
\label{fig:five-outcomes}
\end{figure}

\begin{figure}[t]
\centering
\includegraphics[width=\columnwidth]{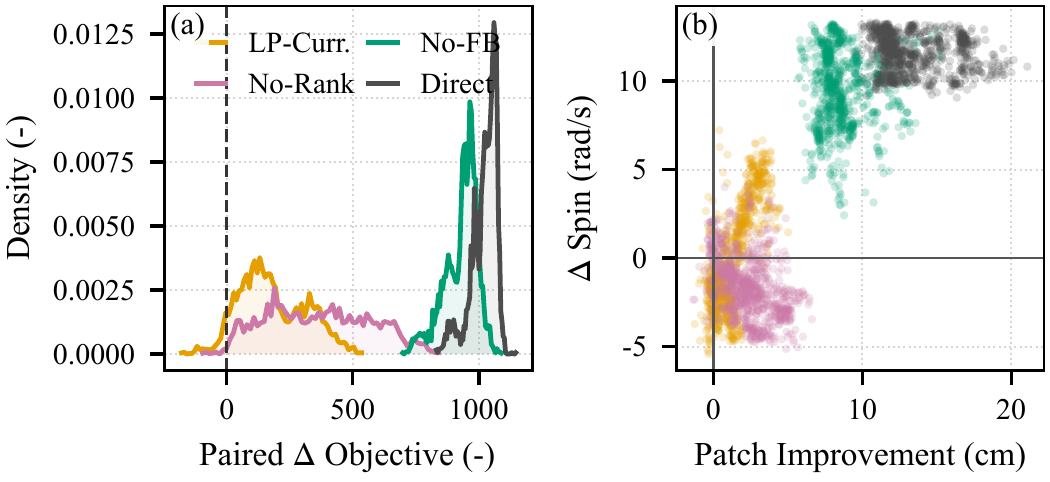}
\caption{Paired comparisons quantify the RISE effect through (a) score differences and (b) patch-precision improvement versus spin change; abbreviations follow Fig.~\ref{fig:five-outcomes}.}
\label{fig:five-paired}
\end{figure}

\begin{figure*}[t]
\centering
\includegraphics[width=\textwidth]{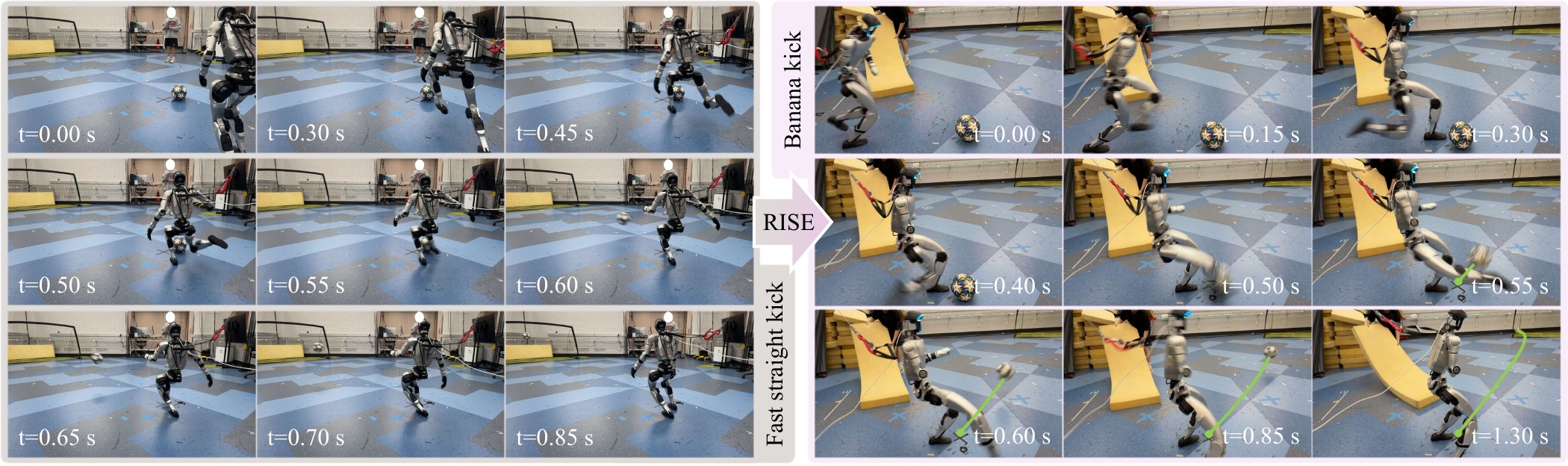}
\caption{Straight- and curved-kick executions show approach, contact, and follow-through, with overlaid paths illustrating their post-contact differences; calibrated flight measurements are reported in Fig.~\ref{fig:flight}.}
\label{fig:real-evolution}
\end{figure*}

\begin{figure}[t]
\centering
\includegraphics[width=\columnwidth]{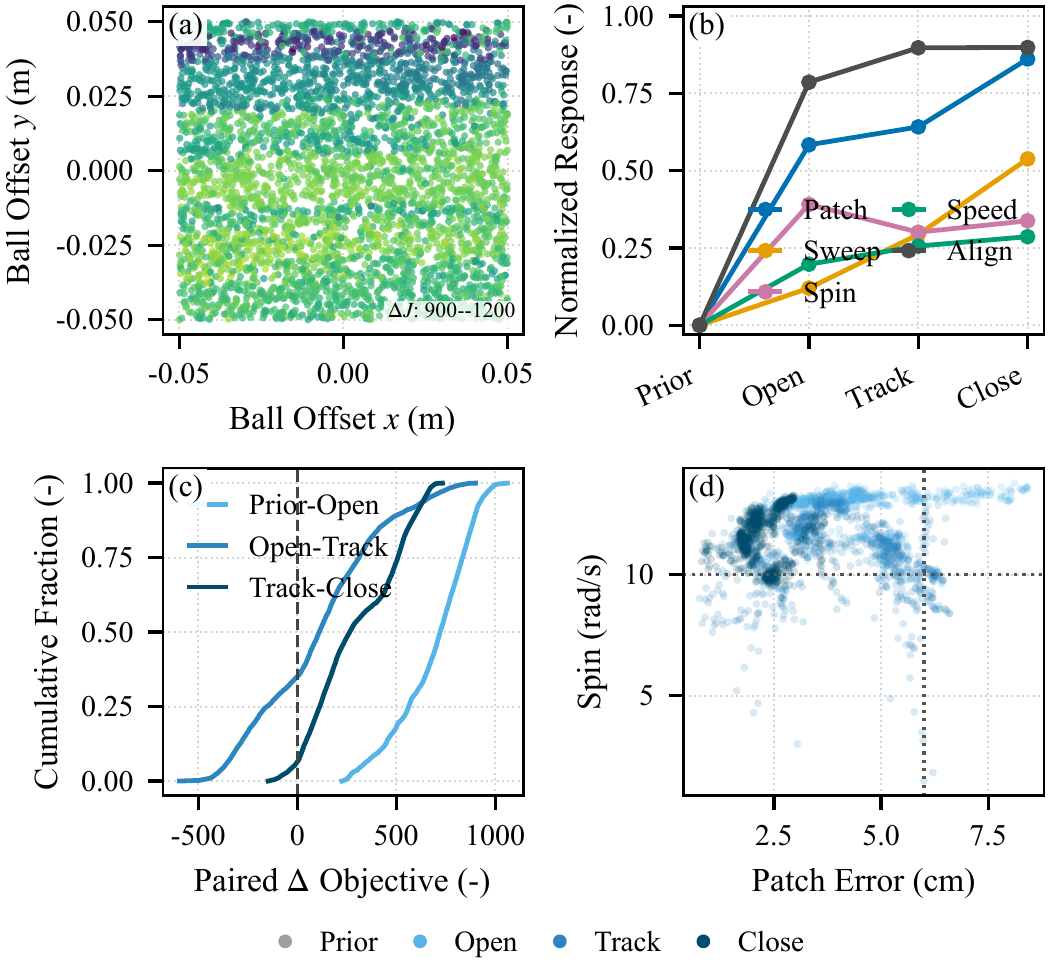}
\caption{Physical-response evolution across accepted RISE checkpoints: (a) prior-to-final gain over ball offsets, (b) normalized response coordinates, (c) consecutive paired-score increments, and (d) patch-error--spin distributions relative to the joint thresholds.}
\label{fig:response-evolution}
\end{figure}

\begin{table*}[t]
\caption{Cross-method comparison over five independent seeds under the fixed evaluation score; brackets denote 95\% matched-context bootstrap intervals.}
\label{tab:matched}
\centering
\setlength{\tabcolsep}{1.5pt}
\begin{tabular*}{\textwidth}{@{\extracolsep{\fill}}lcccccc@{}}
\toprule
Method & Score & Gap $\downarrow$ & Coverage (\%) & $\operatorname{CVaR}_{10}$ & Regret $\downarrow$ & RISE win (\%) \\
\midrule
RISE & 1093.7 [1092.1,1095.3] & 431.3 & 50.88 & 975.5 & 0.0149 & --- \\
LP curriculum & 912.8 [909.3,916.3] & 612.2 & 15.23 & 710.9 & 0.1427 & 94.21 \\
RISE w/o ranking & 729.6 [724.2,735.0] & 795.4 & 0.073 & 414.2 & 0.0846 & 99.34 \\
RISE w/o feedback & 163.2 [162.1,164.3] & 1361.8 & 0.000 & 86.4 & 1.2135 & 100.00 \\
Direct PPO & 71.7 [71.5,72.0] & 1453.3 & 0.000 & 58.1 & 1.9466 & 100.00 \\
\bottomrule
\end{tabular*}
\end{table*}

For the selected objective, PPO \cite{schulman2017ppo} is warm-started from $\theta_j$ to produce a candidate policy $\widetilde\theta$, while $\eta'_j$ and $P_\psi$ remain fixed during the inner update. The changed reward then guides contact adaptation from the accepted checkpoint.

Acceptance is based on normalized physical-response displacement. Let $\mu(\theta)=\mathbb E_{p_\theta}[y]$. The unit vector $u_j$ selects the response coordinates targeted by the candidate and their desired directions, while $S_y=\operatorname{diag}(b_1,\ldots,b_m)$ normalizes their physical units. Using policy means estimated on matched contexts, the response-frontier advance is
\begin{equation}
\begin{aligned}
\Delta_j
=\big\langle
S_y^{-1}\big[\mu(\widetilde\theta)-\mu(\theta_j)\big],
u_j
\big\rangle .
\end{aligned}
\label{eq:frontier}
\end{equation}
This measures directional progress of the mean response. A candidate is accepted if it exceeds the threshold $\kappa>0$ and retains the required reliability:
\begin{equation}
\Delta_j\ge\kappa,\qquad q(\widetilde\theta)\ge q_{\min}.
\label{eq:acceptance}
\end{equation}

Acceptance sets $(\theta_{j+1},\eta_{j+1})=(\widetilde\theta,\eta'_j)$ and expands the radius to $\min(\gamma_\uparrow\delta_j,\delta_{\max})$. Rejection retains $(\theta_j,\eta_j)$ and contracts the radius to $\gamma_\downarrow\delta_j$, after which a new objective is proposed. Here, $\gamma_\uparrow>1$ and $0<\gamma_\downarrow<1$. The loop terminates when the training budget is exhausted, the radius falls below $\delta_{\min}$, or no admissible proposal provides sufficient excitation gain. Let $j_*$ index the final accepted iterate of the adaptation procedure. The final accepted policy $\pi_{\theta_{j_*}}$ is frozen and transferred to the physical robot without further learning. Contact calibration and aerodynamic modeling enter through the training dynamics $P_\psi$; the RISE objective-selection and PPO update loops terminate before deployment. Separate motion-capture recordings of the physical ball flight quantify the transferred response in Section~\ref{sec:physical}.

\section{Results and discussion}
\label{sec:results}

\subsection{Experimental setup}
\label{sec:setup}

Experiments use a 23-joint Unitree G1 and an ordinary-kick prior trained from retargeted mocap references following \cite{he2025asap}. IsaacSim/PhysX runs at 200~Hz and the policy at 50~Hz. PPO uses 4,096 parallel environments, 24-step rollouts, five learning epochs, four minibatches, $\gamma=0.99$, GAE $\lambda=0.95$, and actor/critic MLPs with widths $[512,256,128]$. We compare RISE with a learning-progress curriculum based on automatic-teacher methods \cite{portelas2020teacher} and the TeachMyAgent setting \cite{romac2021teachmyagent}, direct-final PPO, and two ablations removing sensitivity ranking or closed-loop acceptance. All methods use the same environment-interaction budget and five independent random seeds. For each seed, the terminal policy is evaluated using the actor mean on 4,096 matched ball-pose contexts with offsets within $\pm5$ cm. Cross-method results aggregate the five seeds over these matched contexts. The numerical settings are listed in Table~\ref{tab:exp-config}. Fig.~\ref{fig:sim-to-real-visual} summarizes the simulated kick, the straight-versus-banana trajectory difference, and the matched sim-to-real motion comparison. Because rewards change during adaptation, every evaluated trajectory is rescored using one fixed physical-response objective,

\begin{table*}[t]
\caption{Simulated terminal-response diagnostics for the five policies. Spin is in rad/s, patch error in cm, and speeds in m/s.}
\label{tab:physical-outcomes}
\centering
\setlength{\tabcolsep}{1.5pt}
\begin{tabular*}{\textwidth}{@{\extracolsep{\fill}}lccccc@{}}
\toprule
Method & Spin & Patch error $\downarrow$ & Tangential speed & Ball speed & Alignment \\
\midrule
RISE & 11.549 & 2.274 & 8.059 & 8.711 & 0.899 \\
LP curriculum & 11.553 & 3.575 & 7.644 & 8.619 & 0.919 \\
RISE w/o ranking & 13.210 & 4.631 & 6.584 & 8.418 & 0.912 \\
RISE w/o feedback & 1.887 & 11.099 & 6.172 & 10.073 & 0.764 \\
Direct PPO & 0.000 & 15.776 & 4.988 & 8.296 & 0.845 \\
\bottomrule
\end{tabular*}
\end{table*}

\begin{align}
J_{\mathrm{eval}}={}&
\lambda_{p}\phi_{\mathrm{bell}}(e_{\mathrm{patch}};0,s_{p})
+\lambda_{t}\phi_{\mathrm{bell}}(v_{\mathrm{tan}};t_{t},s_{t})\nonumber\\
&+\lambda_{\omega}\phi_{\mathrm{sat}}(\omega;s_{\omega})
+\lambda_{v}\phi_{\mathrm{sat}}(v_{\mathrm{ball}};s_{v}),
\label{eq:common-score}
\end{align}
with $J_{\max}=\lambda_{p}+\lambda_{t}+\lambda_{\omega}+\lambda_{v}$. The normalized contact-alignment response is denoted by $\alpha$. Joint target-set coverage also includes alignment and uses thresholds
$(\bar e,\bar v,\bar\omega,\bar\alpha)$ on patch error, tangential speed, spin, and contact alignment. Define the normalized threshold-deficit vector
$\mathbf z=(e_{\mathrm{patch}}/\bar e-1,\,
1-\omega/\bar\omega,\,
1-v_{\mathrm{tan}}/\bar v,\,
1-\alpha/\bar\alpha)$. The response regret is
\begin{equation}
\mathcal{R}=\|[\mathbf z]_{+}\|_{2},
\label{eq:regret}
\end{equation}
where $[\cdot]_{+}$ is elementwise rectification. We report $J_{\mathrm{eval}}$, the gap $J_{\max}-J_{\mathrm{eval}}$, joint coverage, lower-tail $\operatorname{CVaR}_{10}$ (the mean score in the worst 10\% of contexts), response regret $\mathcal{R}$, and the fraction of matched contexts in which RISE outperforms each comparator. Joint coverage is the fraction with $\mathcal{R}=0$. The evaluation parameters and thresholds are listed in Table~\ref{tab:exp-config}.

\begin{figure*}[t]
\centering
\includegraphics[width=\textwidth]{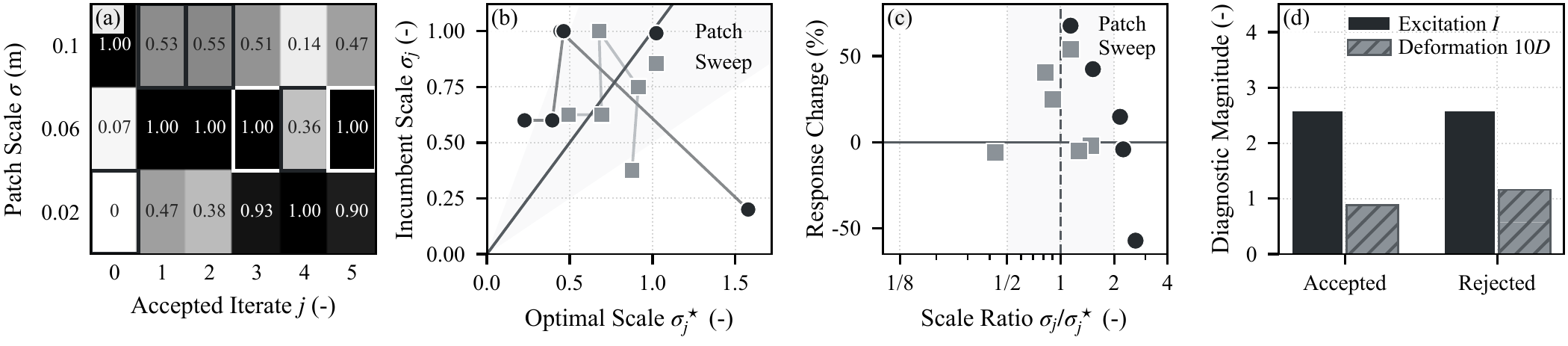}
\caption{Sensitivity-scale diagnostics across accepted RISE checkpoints: (a) patch excitation over admissible scales, (b) incumbent versus response-matched reference scales, (c) scale mismatch versus subsequent channel change, and (d) excitation and deformation for accepted and rejected candidates.}
\label{fig:geom}
\end{figure*}

\begin{figure}[t]
\centering
\includegraphics[width=\columnwidth]{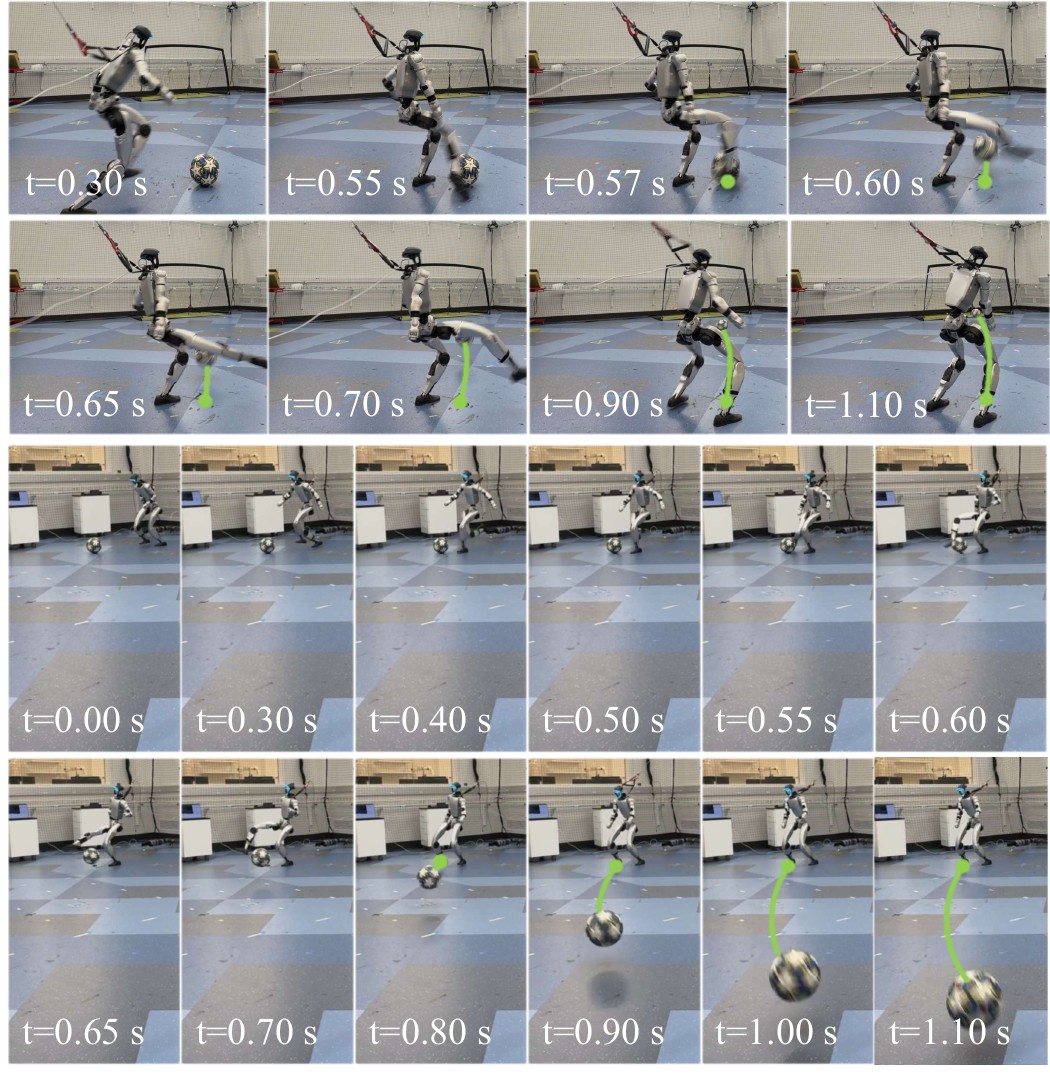}
\caption{Physical curved-kick execution through approach, contact, and follow-through, with overlays illustrating the observed post-contact ball path; quantitative trajectories are reported in Fig.~\ref{fig:flight}.}
\label{fig:real-frames}
\end{figure}

\begin{figure}[t]
\centering
\includegraphics[width=\columnwidth]{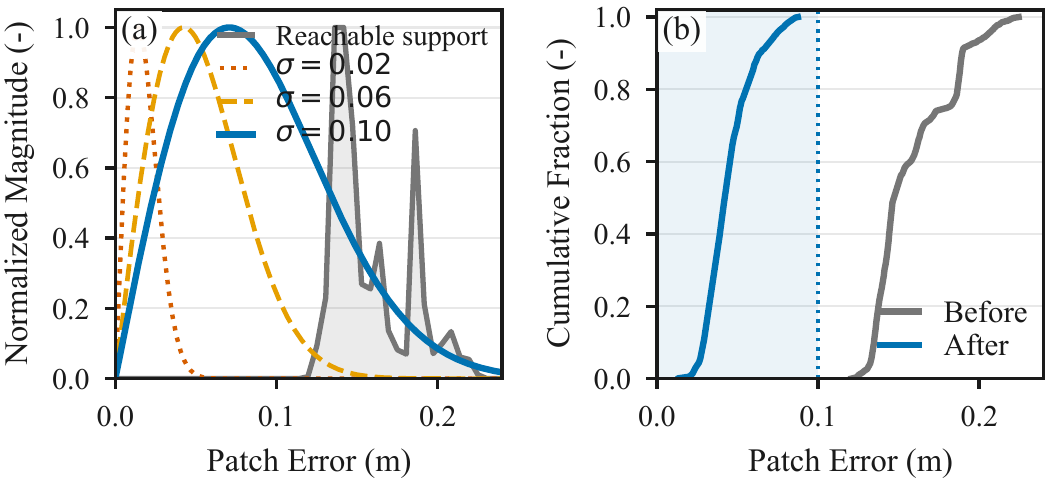}
\caption{Measured response-space starvation and repair: (a) reachable patch-error support versus reward sensitivity under admissible scales and (b) response distributions before and after correction.}
\label{fig:concept}
\end{figure}

\begin{figure}[t]
\centering
\includegraphics[width=\columnwidth]{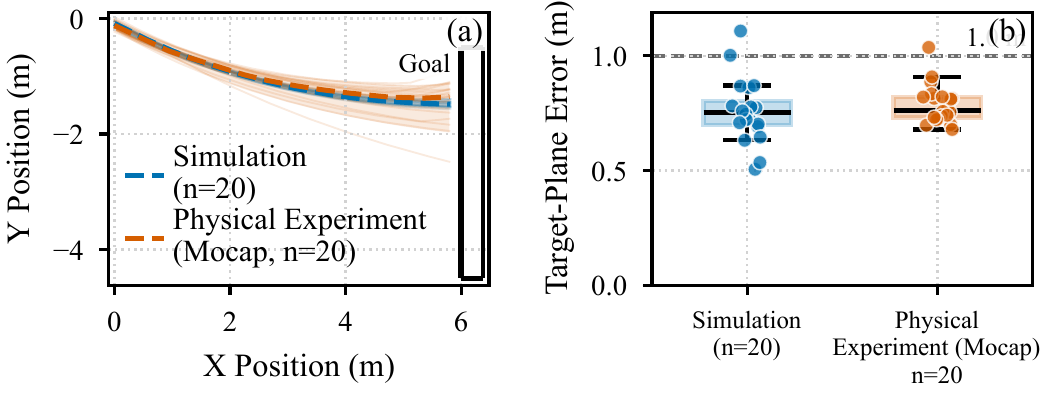}
\caption{Frozen-policy sim-to-real validation: (a) simulated and motion-capture-recorded trajectories; and (b) target-plane errors. Here, 20 simulated and 20 physical trajectories, randomly sampled from the original dataset, are shown for clarity.}
\label{fig:flight}
\end{figure}

\subsection{Cross-method performance and component ablations}
\label{sec:comparison}

Cross-method score and regret distributions are shown in Fig.~\ref{fig:five-outcomes}, while Fig.~\ref{fig:five-paired} resolves the paired RISE effects across matched contexts. Fig.~\ref{fig:real-evolution} qualitatively contrasts the ordinary and curved-kick executions, and the aggregate metrics are summarized in Table~\ref{tab:matched}. RISE attains a mean evaluation score of 1093.7, exceeding the learning-progress curriculum by 19.8\%. Relative to the same comparator, joint target coverage increases from 15.23 to 50.88\% and response regret decreases from 0.1427 to 0.0149. This advantage is not explained by spin alone: the learning-progress curriculum reaches nearly identical mean spin, while removing sensitivity ranking raises spin further to 13.210 rad/s but degrades patch error, tangential speed, and joint target coverage to 0.073\%. RISE instead evolves toward a coordinated contact regime that jointly satisfies spin generation, contact placement, tangential impact, and reliability. Without acceptance feedback, the policy produces the highest ball speed but only 1.887 rad/s spin and 11.099 cm patch error. Direct-final PPO produces zero spin in the terminal simulation evaluation.

\subsection{Response evolution during adaptation}
\label{sec:mechanism}

To examine contact and spin evolution, we track accepted checkpoints from one representative RISE run on the same context bank. The first displayed update raises the mean score from near zero to 688.2, improving every matched context but yielding no joint target coverage. The next checkpoint reaches 792.8 and reduces response regret from 0.367 to 0.105. At the final checkpoint, the score reaches 1093.7, regret falls to 0.0149, and coverage reaches 50.88\%, showing that continuous response measures capture progress before all target thresholds are met (Fig.~\ref{fig:response-evolution}).

At initialization, mean patch error is 15.78 cm with a 2 cm patch-kernel scale, giving raw excitation of $2.07\times10^{-16}$, while spin is zero across all evaluation contexts. The first accepted patch-scale update widens the kernel from 0.02 to 0.10 m; patch error then drops to 4.45 cm and spin rises to 12.671 rad/s. This sequence supports Corollary~\ref{cor:sat}: adapting a coupled contact response first moves the policy beyond the prior's contact regime, after which substantial spin emerges despite the initially flat spin channel.

Archived objective edits provide a retrospective check of the sensitivity criterion (Fig.~\ref{fig:geom}). For bell-kernel scale edits, the excitation-maximizing candidate matches the recorded choice in four of five transitions, with the remaining choice differing by only 2\% in normalized excitation. We use Fig.~\ref{fig:geom} as a mechanism-consistency diagnostic rather than an independent benchmark. The physical execution is shown in Fig.~\ref{fig:real-frames}, while Fig.~\ref{fig:concept} visualizes response-space starvation and repair and Table~\ref{tab:physical-outcomes} summarizes terminal response diagnostics.

\subsection{Physical deployment}
\label{sec:physical}

For hardware transfer, the final RISE policy is frozen and deployed without further learning. We evaluate 60 simulated and 30 motion-capture-recorded physical trials. Fig.~\ref{fig:flight} displays 20 simulated and 20 physical trajectories for clarity, while all reported physical statistics use the full set of 30 physical trials. Simulated flights bend consistently in one lateral direction, with a median top-view bow of 0.381 m and a median heading change of 33.1 degrees; the physical trials exhibit the same bending direction. The curved-flight behavior is preserved under direct sim-to-real transfer without real-world policy adaptation.

\section{Conclusions}
\label{sec:conclusion}

This work demonstrates skill evolution beyond performance refinement in a physical humanoid: starting from a reliable ordinary-kick prior, RISE drives the policy into a qualitatively different spin-generating contact regime that produces repeatable curved ball flight after sim-to-real transfer. More broadly, the results show that a strong motion prior need not be a terminal skill template, and it can serve as a scaffold from which qualitatively new contact-rich behavior emerges. The contributions of this work are summarized as follows:

\begin{itemize}
    
    \item RISE evolves the learning objective with the policy's reachable physical responses and retains updates only after verified response progress and kicking reliability, allowing a competent prior to leave its familiar contact regime rather than merely refine speed or accuracy.

    \item The response-space analysis characterizes first-order learning starvation and explains why coupled contact responses can restore a viable learning direction when the target response itself is locally uninformative.
    
    \item In simulation and experiment, RISE transforms an ordinary kick into a high-spin banana kick with more than 11.55~rad/s mean ball spin, improves the mean evaluation score by 19.8\% over a learning-progress curriculum, and raises joint target attainment from 15.2\% to 50.9\%. Repeated physical trials reproduce the learned curved-kick behavior on real hardware without further fine-tuning.

\end{itemize}

\balance
\bibliographystyle{IEEEtran}
\bibliography{references}

@book{allgower1990,
  title={Numerical Continuation Methods: An Introduction},
  author={Allgower, Eugene L. and Georg, Kurt},
  series={Springer Series in Computational Mathematics},
  volume={13},
  publisher={Springer},
  year={1990},
  doi={10.1007/978-3-642-61257-2}
}

@article{kiefer1959optimal,
  title={Optimum Experimental Designs},
  author={Kiefer, Jack},
  journal={Journal of the Royal Statistical Society: Series B (Methodological)},
  volume={21},
  number={2},
  pages={272--304},
  year={1959},
  doi={10.1111/j.2517-6161.1959.tb00338.x}
}

@article{box1961fractional,
  title={The $2^{k-p}$ Fractional Factorial Designs, Part I},
  author={Box, George E. P. and Hunter, J. Stuart},
  journal={Technometrics},
  volume={3},
  number={3},
  pages={311--351},
  year={1961},
  doi={10.1080/00401706.1961.10489951}
}

@article{wang2016rembo,
  title={Bayesian Optimization in a Billion Dimensions via Random Embeddings},
  author={Wang, Ziyu and Hutter, Frank and Zoghi, Masrour and Matheson, David and de Freitas, Nando},
  journal={Journal of Artificial Intelligence Research},
  volume={55},
  pages={361--387},
  year={2016},
  doi={10.1613/jair.4806}
}

@inproceedings{kandasamy2015additive,
  title={High Dimensional Bayesian Optimisation and Bandits via Additive Models},
  author={Kandasamy, Kirthevasan and Schneider, Jeff and P{\'o}czos, Barnab{\'a}s},
  booktitle={Proceedings of the 32nd International Conference on Machine Learning},
  series={Proceedings of Machine Learning Research},
  volume={37},
  pages={295--304},
  year={2015}
}

@inproceedings{eriksson2021saasbo,
  title={High-Dimensional Bayesian Optimization with Sparse Axis-Aligned Subspaces},
  author={Eriksson, David and Jankowiak, Martin},
  booktitle={Proceedings of the Thirty-Seventh Conference on Uncertainty in Artificial Intelligence},
  series={Proceedings of Machine Learning Research},
  volume={161},
  pages={493--503},
  year={2021}
}

@inproceedings{papenmeier2022baxus,
  title={Increasing the Scope as You Learn: Adaptive Bayesian Optimization in Nested Subspaces},
  author={Papenmeier, Leonard and Nardi, Luigi and Poloczek, Matthias},
  booktitle={Advances in Neural Information Processing Systems},
  volume={35},
  year={2022}
}

@inproceedings{pathak2019disagreement,
  title={Self-Supervised Exploration via Disagreement},
  author={Pathak, Deepak and Gandhi, Dhiraj and Gupta, Abhinav},
  booktitle={Proceedings of the 36th International Conference on Machine Learning},
  series={Proceedings of Machine Learning Research},
  volume={97},
  pages={5062--5071},
  year={2019}
}

@inproceedings{eysenbach2019diayn,
  title={Diversity Is All You Need: Learning Skills without a Reward Function},
  author={Eysenbach, Benjamin and Gupta, Abhishek and Ibarz, Julian and Levine, Sergey},
  booktitle={International Conference on Learning Representations},
  year={2019}
}

@inproceedings{sharma2020dads,
  title={Dynamics-Aware Unsupervised Discovery of Skills},
  author={Sharma, Archit and Gu, Shixiang and Levine, Sergey and Kumar, Vikash and Hausman, Karol},
  booktitle={International Conference on Learning Representations},
  year={2020}
}

@article{pugh2016quality,
  title={Quality Diversity: A New Frontier for Evolutionary Computation},
  author={Pugh, Justin K. and Soros, Lisa B. and Stanley, Kenneth O.},
  journal={Frontiers in Robotics and AI},
  volume={3},
  pages={40},
  year={2016},
  doi={10.3389/frobt.2016.00040}
}

@inproceedings{johannink2019residual,
  title={Residual Reinforcement Learning for Robot Control},
  author={Johannink, Tobias and Bahl, Shikhar and Nair, Ashvin and Luo, Jianlan and Kumar, Avinash and Loskyll, Matthias and Ojea, Juan Aparicio and Solowjow, Eugen and Levine, Sergey},
  booktitle={2019 International Conference on Robotics and Automation},
  pages={6023--6029},
  year={2019},
  doi={10.1109/ICRA.2019.8794127}
}

@inproceedings{rana2023residual,
  title={Residual Skill Policies: Learning an Adaptable Skill-based Action Space for Reinforcement Learning for Robotics},
  author={Rana, Krishan and Xu, Ming and Tidd, Brendan and Milford, Michael and S{\"u}nderhauf, Niko},
  booktitle={Proceedings of The 6th Conference on Robot Learning},
  series={Proceedings of Machine Learning Research},
  volume={205},
  pages={2095--2104},
  year={2023}
}

@inproceedings{bengio2009curriculum,
  title={Curriculum Learning},
  author={Bengio, Yoshua and Louradour, J{\'e}r{\^o}me and Collobert, Ronan and Weston, Jason},
  booktitle={Proceedings of the 26th International Conference on Machine Learning},
  pages={41--48},
  year={2009},
  doi={10.1145/1553374.1553380}
}

@inproceedings{ng1999policy,
  title={Policy Invariance under Reward Transformations: Theory and Application to Reward Shaping},
  author={Ng, Andrew Y. and Harada, Daishi and Russell, Stuart},
  booktitle={Proceedings of the 16th International Conference on Machine Learning},
  pages={278--287},
  year={1999}
}

@article{schulman2017ppo,
  title={Proximal Policy Optimization Algorithms},
  author={Schulman, John and Wolski, Filip and Dhariwal, Prafulla and Radford, Alec and Klimov, Oleg},
  journal={arXiv preprint arXiv:1707.06347},
  year={2017}
}

@article{peng2018deepmimic,
  title={DeepMimic: Example-Guided Deep Reinforcement Learning of Physics-Based Character Skills},
  author={Peng, Xue Bin and Abbeel, Pieter and Levine, Sergey and van de Panne, Michiel},
  journal={ACM Transactions on Graphics},
  volume={37},
  number={4},
  pages={143:1--143:14},
  year={2018},
  doi={10.1145/3197517.3201311}
}

@article{he2025asap,
  title={{ASAP}: Aligning Simulation and Real-World Physics for Learning Agile Humanoid Whole-Body Skills},
  author={He, Tairan and Gao, Jiawei and Xiao, Wenli and Zhang, Yuanhang and Wang, Zi and Wang, Jiashun and Luo, Zhengyi and He, Guanqi and Sobanbabu, Nikhil and Pan, Chaoyi and Yi, Zeji and Qu, Guannan and Kitani, Kris and Hodgins, Jessica and Fan, Linxi Jim and Zhu, Yuke and Liu, Changliu and Shi, Guanya},
  journal={Robotics: Science and Systems},
  year={2025}
}

@article{haarnoja2023soccer,
  title={Learning Agile Soccer Skills for a Bipedal Robot with Deep Reinforcement Learning},
  author={Haarnoja, Tuomas and Moran, Ben and Lever, Guy and Huang, Sandy H. and Tirumala, Dhruva and Humplik, Jan and Wulfmeier, Markus and Tunyasuvunakool, Saran and Siegel, Noah Y. and Hafner, Roland and Bloesch, Michael and Hartikainen, Kristian and Byravan, Arunkumar and Hasenclever, Leonard and Tassa, Yuval and Sadeghi, Fereshteh and Batchelor, Nathan and Casarini, Federico and Saliceti, Simone and Game, Claudio and Sreendra, Neel and Patel, Kushal and Gwira, Mani and Huber, Andrea and Hurley, Nora and Nori, Francesco and Hadsell, Raia and Heess, Nicolas},
  journal={Science Robotics},
  volume={9},
  number={89},
  year={2024},
  doi={10.1126/scirobotics.adi8022}
}

@inproceedings{portelas2020teacher,
  title={Teacher Algorithms for Curriculum Learning of Deep {RL} in Continuously Parameterized Environments},
  author={Portelas, R\'{e}my and Colas, C\'{e}dric and Hofmann, Katja and Oudeyer, Pierre-Yves},
  booktitle={Proceedings of the Conference on Robot Learning},
  series={Proceedings of Machine Learning Research},
  volume={100},
  pages={835--853},
  year={2020}
}

@inproceedings{romac2021teachmyagent,
  title={{TeachMyAgent}: A Benchmark for Automatic Curriculum Learning in Deep {RL}},
  author={Romac, Cl\'{e}ment and Portelas, R\'{e}my and Colas, C\'{e}dric and Hofmann, Katja and Oudeyer, Pierre-Yves},
  booktitle={Proceedings of the 38th International Conference on Machine Learning},
  series={Proceedings of Machine Learning Research},
  volume={139},
  pages={9052--9063},
  year={2021}
}

@article{zhang2025multi,
  title={Multi-scale reinforcement learning of dynamic energy controller for connected electrified vehicles},
  author={Zhang, Hao and Lei, Nuo and Li, Shengbo Eben and Zhang, Junzhi and Wang, Zhi},
  journal={IEEE Transactions on Intelligent Transportation Systems},
  year={2025},
  publisher={IEEE}
}

@article{zhang2025bi,
  title={Bi-level transfer learning for lifelong-intelligent energy management of electric vehicles},
  author={Zhang, Hao and Lei, Nuo and Peng, Wang and Li, Bingbing and Lv, Shujun and Chen, Boli and Wang, Zhi},
  journal={IEEE Transactions on Intelligent Transportation Systems},
  volume={26},
  number={10},
  pages={16174--16187},
  year={2025},
  publisher={IEEE}
}

@article{zhang2026cognition,
  title={Cognition to Control-Multi-Agent Learning for Human-Humanoid Collaborative Transport},
  author={Zhang, Hao and Zhao, Ding and Tseng, H Eric},
  journal={arXiv preprint arXiv:2603.03768},
  year={2026}
}

@article{zhang2026interaction,
  title={Interaction-Aware Whole-Body Control for Compliant Object Transport},
  author={Zhang, Hao and Tseng, Yves and Zhao, Ding and Tseng, H Eric},
  journal={arXiv preprint arXiv:2603.03751},
  year={2026}
}

@article{zhang2026halo,
  title={HALO: Learning Human-Robot Collaboration via Heterogeneous-Agent Lyapunov Policy Optimization},
  author={Zhang, Hao and Niu, Yaru and Wang, Yikai and Zhao, Ding and Tseng, H Eric},
  journal={arXiv preprint arXiv:2603.03741},
  year={2026},
  publisher={International Conference on Machine Learning (ICML)}
}

\end{document}